\documentclass[lettersize,journal]{IEEEtran}
\usepackage{amsmath,amsfonts}
\usepackage{algorithmic}
\usepackage{algorithm}
\usepackage{array}
\usepackage{textcomp}
\usepackage{stfloats}
\usepackage{url}
\usepackage{verbatim}
\usepackage{graphicx}
\usepackage{cite}
\usepackage{booktabs}
\usepackage{multirow}
\usepackage{graphicx}
\usepackage[caption=false,labelformat=simple]{subfig}

\begin{document}
\title{Scale-Aware Neural Architecture Search for Multivariate Time Series Forecasting}

\author{Donghui Chen, Ling Chen,  Zongjiang Shang, Youdong Zhang, Bo Wen, and Chenghu Yang
\thanks{This work was supported by the National Key Research and Development Program of China under Grant 2018YFB0505000. (Corresponding author: Ling Chen.)}       
\thanks{Donghui Chen, Ling Chen, and Zongjiang Shang are with the College of Computer Science and Technology, Zhejiang University, Hangzhou 310027, China (e-mail: \{chendonghui, lingchen, zongjiangshang\}@cs.zju.edu.cn).}
\thanks{Youdong Zhang, Bo Wen, and Chenghu Yang are with Alibaba Group, Hangzhou 311100, China (e-mail: \{linqing.zyd, wenbo.wb\}@alibaba-inc.com; yexiang.ych@taobao.com).}}

\markboth{ IEEE TRANSACTIONS ON KNOWLEDGE AND DATA ENGINEERING,~Vol.~xx, No.~xx, xx~2021}%
{Shell \MakeLowercase{\textit{et al.}}: A Sample Article Using IEEEtran.cls for IEEE Journals}

\def\mathbi#1{\textbf{\em #1}}
\maketitle

\begin{abstract}
Multivariate time series (MTS) forecasting has attracted much attention in many intelligent applications. It is not a trivial task, as we need to consider both intra-variable dependencies and inter-variable dependencies. However, existing works are designed for specific scenarios, and require much domain knowledge and expert efforts, which is difficult to transfer between different scenarios. In this paper, we propose a scale-aware neural architecture search framework for MTS forecasting (SNAS4MTF). A multi-scale decomposition module transforms raw time series into multi-scale sub-series, which can preserve multi-scale temporal patterns. An adaptive graph learning module infers the different inter-variable dependencies under different time scales without any prior knowledge. For MTS forecasting, a search space is designed to capture both intra-variable dependencies and inter-variable dependencies at each time scale. The multi-scale decomposition, adaptive graph learning, and neural architecture search modules are jointly learned in an end-to-end framework. Extensive experiments on two real-world datasets demonstrate that SNAS4MTF achieves a promising performance compared with the state-of-the-art methods. 
\end{abstract}

\begin{IEEEkeywords}
Multivariate time series forecasting, neural architecture search, graph learning, multi-scale decomposition.
\end{IEEEkeywords}

\section{Introduction}
\IEEEPARstart{W}{ith} the rapid development of sensing and communication technology in Cyber-Physical Systems (CPS), multivariate time series (MTS) are ubiquitous. MTS forecasting, which aims at forecasting the future values based on a group of observed time series in the past, has been widely studied in recent years \cite{ref1,ref2,ref3,ref4,ref5}. It has played important roles in many applications, e.g., a better driving route can be planned based on the forecasted traffic flows, and an investment strategy can be designed with the forecasting of the near-future stock market.

Making accurate MTS forecasting is a challenging task, as both intra-variable dependencies (i.e., the temporal dependencies within one time series) and inter-variable dependencies (i.e., the forecasting values of a single variable are affected by the values of other variables) need to be considered. To solve this problem, many classical network architectures have been designed. Temporal convolutional network, (dilated) recurrent neural network, (dilated) convolutional neural network, (sparse) self-attention mechanism, and self-attention mechanism with convolution are exploited to model intra-variable dependencies \cite{ref6,ref7,ref8,ref9,ref10}. Convolutional neural network, (multi-hop) graph neural network, and the latest graph learning based models are exploited to model inter-variable dependencies \cite{ref11,ref12,ref13,ref14}. By stacking or integrating multiple these networks to model intra-variable dependencies and inter-variable dependencies jointly, existing methods have achieved promising results \cite{ref15,ref16}. However, existing methods are usually designed for specific scenarios or specific data distributions, and are difficult to transfer between different scenarios, e.g., for one scenario, we should pay more attention to modeling intra-variable dependencies, while we should pay more attention to modeling inter-variable dependencies for another scenario. Therefore, it is necessary to design a specific network architecture for a scenario, which requires a lot of human intelligence and domain knowledge.

Neural architecture search (NAS) \cite{ref17,ref18}, which defines the search space, the search strategy, and the performance estimation strategy, can automatically find the optimal network architecture for a given scenario. The network architectures discovered by NAS algorithms have outperformed the handcrafted ones at many domains. Motived by the success of NAS, we extend NAS studies to MTS forecasting. However, designing NAS algorithms to find optimal network architectures for MTS forecasting is non-trivial due to the following two challenges.

First, the inter-variable dependencies may be different under different time scales. Existing MTS often appear multi-scale temporal patterns (e.g., daily and weekly). Temporal patterns of different time scales are often affected by different inter-variable dependencies, i.e., we should distinguish the inter-variable dependencies when modeling temporal patterns of different time scales. For example, in the forecasting of the power consumptions of a household, when modeling the short-term patterns, it might be essential to pay more attention to the power consumptions of its neighbors, e.g., a community party will affect the power consumption of some households in the community. When modeling the long-term patterns, it might be essential to pay more attention to the households that have similar living habits, e.g., the daily and weekly temporal patterns would be similar in these households that have the same working and sleeping hours. Existing MTS forecasting methods \cite{ref12,ref13,ref14,ref15,ref16,guo2021learning,ref19} exploit extra information (e.g., road networks and physical structures) or learn an adjacency matrix to denote the inter-variable dependencies, which is hard to describe the different inter-variable dependencies under different time scales, and makes the models biased to learn one type of prominent and shared temporal patterns.

Second, the search space of MTS forecasting is different with the ones in existing NAS works. Existing NAS works are mainly designed for modeling the grid data (e.g., images), which fixes the search space and focuses on designing better search strategies \cite{ref20,ref21}. These works define the convolutional neural networks as the basic operation blocks in the search space and choose optimal kernel size for the convolution operations. In contrast, for MTS forecasting, the basic operation blocks in the search space should effectively capture both intra-variable dependencies and inter-variable dependencies. Pan et al. \cite{ref20} first proposed a NAS method for modeling spatio-temporal graphs and designed a search space, which includes temporal convolution and spatial convolution operations. However, this work exploits additional information (e.g., node attributes and edge attributes) to obtain the adjacency matrix required for the spatial convolution operation, which is not suitable for pure MTS forecasting. In addition, this work constructs the search space by stacking multiple identical spatio-temporal modules, which makes the models prefer to learn one type of prominent and shared temporal patterns, and fails to model multi-scale patterns.

To tackle the abovementioned problems, we propose a scale-aware neural architecture search framework for multivariate time series forecasting (SNAS4MTF). It integrates multi-scale decomposition and adaptive graph learning into the NAS framework, and automatically captures both intra-variable dependencies and inter-variable dependencies. In summary, our contributions are as follows:

\begin{itemize}
\item{Construct a search space for MTS forecasting, which includes temporal convolution, graph convolution, temporal attention, spatial attention, and other basic operations. By decomposing the time series into multi-scale sub-series, SNAS4MTF searches an optimal network architecture for each time scale and learn multi-scale temporal patterns.}
\item{Introduce an adaptive graph learning (AGL) module to learn adjacent matrices, which can reflect the inter-variable dependencies under different time scales. AGL introduces a scale shared graph learner and matrix decomposition technique to reduce the number of parameters.}
\item{Conduct extensive experiments on two real-world MTS benchmark datasets. Experimental results demonstrate that the performance of SNAS4MTF is better than that of the state-of-the-art methods.}
\end{itemize}

The rest of this paper is organized as follows: Section \uppercase\expandafter{\romannumeral2} reports a review of the related work; Section \uppercase\expandafter{\romannumeral3} gives the details of SNAS4MTF; Section \uppercase\expandafter{\romannumeral4} presents the experimental results; Section \uppercase\expandafter{\romannumeral5} concludes the paper.

\section{Related Work}
\subsection{Multivariate Time Series Forecasting}
Multivariate time series forecasting has been studied intensively for decades. Recently, deep learning-based methods have achieved impressive performances. Early works process the MTS input as vectors and assume that the forecasting values of a single variable are affected by the values of all other variables. For example, Lai et al. \cite{ref2} designed a convolutional neural network to capture short-term temporal dependencies and a recurrent-skip network to capture long-term temporal dependencies. Shih et al. \cite{ref3} exploited attention mechanism to focus on the important input variables and time steps. Chang et al. \cite{ref22} exploited memory network and three separate encoders to capture temporal dependencies. Sen et al. \cite{ref23} proposed a hybrid model that combines a global regularized matrix factorization to derive factors and a temporal network to capture local properties. However, these works assume that each variable is affected by all other variables, which is unreasonable in realistic applications, and may lead to overfitting when the number of variables is large.

Graph based methods \cite{ref24} treat the variables in MTS as the nodes, while the pairwise inter-variable dependencies as edges in the graph, which can explicitly express the inter-variable dependencies according to the existence or the weights of edges. These works model MTS by graph neural network (GNN) to exploit the rich structural information of a graph, and RNN or TCN to capture temporal dependencies. One of the challenges of the graph based MTS forecasting is to obtain a well-defined graph structure (i.e., adjacency matrix) as the inter-variable dependencies. Prior works \cite{ref25,ref26,ref27,ref28,ref29,ref30} exploit extra information (e.g., road networks, physical structures, and extra feature matrices) in their specific scenarios, or non-parametric methods to generate the adjacency matrix. For example, in traffic flow forecasting \cite{ref25,ref27}, the adjacency matrix can be generated through the road network in the city. If there is a connected road between two nodes, an edge is constructed in the graph structure. In ride-hailing demand forecasting \cite{ref31}, three kinds of adjacency matrices are generated: the neighborhood graph to encode the spatial proximity, the functional similarity graph to encode the similarity of surrounding point of interests of regions, and the transportation connectivity graph to encode the connectivity between distant regions. Xu et al. \cite{ref29} calculated the pairwise transfer entropy between variables to construct the adjacency matrix. He et al. \cite{ref30} exploited dynamic time warping (DTW) algorithm to calculate the pairwise pattern similarities, which are regarded as the adjacency matrix. However, these works are limited by the power of similarity function or require a lot of domain knowledge and expert efforts.

Recently, several graph learning based methods \cite{ref1,ref4,ref32,ref33} attempt to directly learn a adjacency matrix from the MTS. Kipf et al. \cite{ref32} first introduced the neural relational inference model that takes the original time series as input and learns the adjacency matrix by variational inference. Graber et al.\cite{ref33} proposed a dynamic neural relational inference model to learn an individual adjacency matrix for each time step. Tang et al. \cite{ref34} and Zheng et al. \cite{ref14} exploited attention mechanisms to learn the pairwise correlation between nodes as the adjacency matrix. In addition, several works \cite{ref1,ref2,ref3,ref4} use a more direct way, i.e., randomly initializing the embeddings of all nodes and computing the similarities between these nodes, to construct the adjacency matrix. The node embeddings can be optimized like other parameters in the neural networks. For example, Wu et al. \cite{ref4} introduced a graph learning module to learn inter-variable dependencies, and stack the GNN and dilated convolution networks. Bai et al. \cite{ref1} introduced a graph learning module to adaptively infer the inter-variable dependencies and exploit a node adaptive parameter learning module to capture node-specific features. However, existing works only learn one adjacency matrix, which is hard to describe the different inter-variable dependencies under different time scales, and makes the models biased to learn one type of prominent and shared temporal patterns.

\subsection{Neural Architecture Search}
Neural architecture search (NAS) \cite{ref17,ref18} has enabled remarkable progress over the last few years, and has been successfully applied to a variety of tasks, e.g., object detection, image recognition, and natural language processing. Early NAS works \cite{ref35,ref36} exploit the reinforcement learning to search the optimal architecture. However, these works consume substantial computational resources. To reduce the computational cost of search process, gradient-based methods \cite{ref21} relax the categorical design choices to continuous variables and then leverage the efficient gradient backpropagation. Recently, several works introduce weight-sharing to further reduce the computational cost of search process \cite{ref37}.

The most closely related study to our work is the AutoSTG method \cite{ref19}, which also incorporates NAS to search the optimal architecture for spatio-temporal graphs. AutoSTG adopts graph convolution and temporal convolution operations in the search space to capture complex spatio-temporal correlations, and employs the meta learning technique to learn the adjacency matrices of graph convolution and the kernels of temporal convolution. However, AutoSTG exploits additional information (i.e., node attributes and edge attributes) to obtain the adjacency matrix, which is not suitable for pure MTS forecasting. In addition, AutoSTG constructs the network architecture by stacking multiple identical spatio-temporal modules, which makes the models prefer to learn one type of prominent and shared temporal patterns, and fails to model multi-scale patterns.

\section{Methodology}
\subsection{Problem Formulation}
{\bf{Problem Statement.}} In this paper, we focus on multi-step MTS forecasting task. Formally, given a sequence of observed time series signals $\boldsymbol{X}=\left[\boldsymbol{x}_1,\boldsymbol{x}_2,...,\boldsymbol{x}_t,...,\boldsymbol{x}_T \right]\in\mathbb{R}^{N{\times}T}$, where $\boldsymbol{x}_t\in\mathbb{R}^N$ denotes the values at time step $t$, $N$ is the variable dimension, and $\boldsymbol{x}_{t,i}$ denotes the value of the $i^{\text{th}}$ variable at time step $t$, multi-step MTS forecasting task aims at forecasting the future values $\left[\boldsymbol{\widehat{x}}_{T+1},\boldsymbol{\widehat{x}}_{T+2},...,\boldsymbol{\widehat{x}}_{T+h}\right]\in\mathbb{R}^{N{\times}h}$ between time step $T+1$ and $T+h$, where $h$ denotes the look-ahead horizon. The problem can be formulated as:  
\begin{equation}
\label{1}
\left[\boldsymbol{\widehat{x}}_{T+1},\boldsymbol{\widehat{x}}_{T+2},...,\boldsymbol{\widehat{x}}_{T+h}\right]=\mathcal{F}\left(\boldsymbol{x}_1,\boldsymbol{x}_2,...,\boldsymbol{x}_T;\theta\right),
\end{equation}
where $\mathcal{F}$ is the mapping function and $\theta$ denotes all learnable parameters.

{\bf{Definition 1.}} $\emph{MTS to Graph.}$ A graph is defined as $\mathcal{G}=\left(\mathcal{V},\boldsymbol{A}\right)$, where $\mathcal{V}$ denotes the node set and $\left|\mathcal{V}\right|=N$. ${\boldsymbol{A}\in\mathbb{R}^{N{\times}N}}$ is the weighted adjacency matrix, where $\boldsymbol{A}_{i,j}>0$ indicates that there is an edge connecting nodes $i$ and $j$. Given the MTS $\boldsymbol{X}\in\mathbb{R}^{N{\times}T}$, the $i^{\text{th}}$ variable is regarded as the  $i^{\text{th}}$ node $v_i\in\mathcal{V},$ the values of $\{\boldsymbol{x}_{1,i},\boldsymbol{x}_{2,i},...,\boldsymbol{x}_{T,i}\}$ are the features of $v_i$, and $\boldsymbol{A}$ describes the inter-variable dependencies of MTS. Then, multi-step MTS forecasting is modified as:
\begin{equation}
\label{2}
\left[\boldsymbol{\widehat{x}}_{T+1},\boldsymbol{\widehat{x}}_{T+2},...,\boldsymbol{\widehat{x}}_{T+h}\right]=\mathcal{F}\left(\boldsymbol{x}_1,\boldsymbol{x}_2,...,\boldsymbol{x}_T;\mathcal{G};\theta\right),
\end{equation}
where $\theta$ denotes all learnable parameters. We focus on pure MTS forecasting task without any prior knowledge, and the weighted adjacency matrix of the graph will be learned.

{\bf{Definition 2.}} $\emph{NAS to MTS Forecasting. }$Given the MTS forecasting dataset. We aim to learn a neural architecture from the training dataset $\mathbb{D}_{\text{train}}$, and achieve the minimum loss on the validation dataset $\mathbb{D}_\text{val}$:
\begin{equation}
\label{3}
\text{min}_{\theta_\text{a}}\mathcal{L}\left(\theta_\text{w}^*\left(\theta_\text{a}\right),\theta_\text{a},\mathbb{D}_\text{val}\right),
\end{equation}
s.t.:
\begin{equation}
\label{4}
\theta_\text{w}^*\left(\theta_\text{a}\right)=\text{arg}\mathop{\text{min}}\limits_{\theta_\text{w}}\mathcal{L}\left(\theta_\text{w},\theta_\text{a},\mathbb{D}_\text{train}\right),
\end{equation}
where $\mathcal{L}\left(\cdot\right)$ is the loss function, $\theta_\text{w}$ and $\theta_\text{a}$ are the weight parameters of networks and architecture parameters, respectively. $\theta_\text{w}^*\left(\theta_\text{a}\right)$ denotes the values of weight parameters when achieving the minimum loss.

\subsection{Framework}
\begin{figure*}[!t]
\centering
\includegraphics[width=5in]{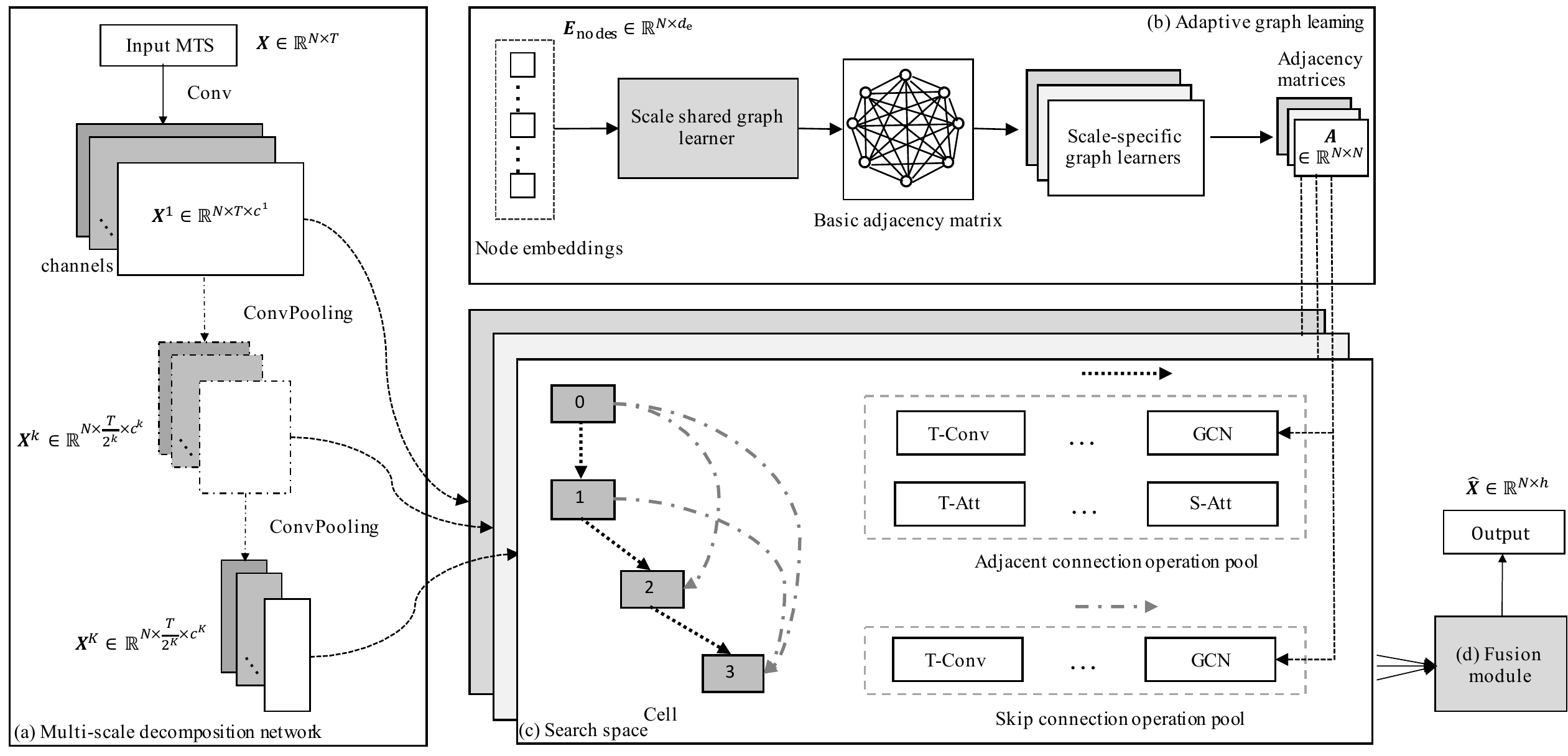}
\caption{The framework of SNAS4MTF.}
\label{Figure_1}
\end{figure*}

Figure \ref{Figure_1} illustrates the framework of SNAS4MTF, which consists of four main parts: a) A multi-scale decomposition network transforms raw time series into multi-scale sub-series, which can preserve the underlying temporal patterns at different time scales. b) An adaptive graph learning module automatically infers inter-variable dependencies under different time scales. It introduces two sub-modules, i.e., a scale shared graph learner and scale-specific graph learners, to reduce the number of parameters. c) A search space is designed for MTS forecasting task, which includes temporal convolution, graph convolution, temporal attention, spatial attention, and other basic operations to automatically capture various inter-variable and intra-variable dependencies. d) A multi-scale fusion and output module obtains the forecasted values, which can effectively promote the feature fusion across different time scales.

\subsection{Multi-Scale Decomposition Network}
A multi-scale decomposition network is designed to preserve the underlying temporal dependencies at different time scales. It exploits multiple decomposition layers to hierarchically transform time series from smaller scale to larger scale. Such multi-scale structure gives us the opportunity to observe raw time series under different time scales \cite{ref5,ref38}. Specifically, the smaller scale feature representations can retain more fine-grained details, while the larger scale feature representations can capture the slow-varying trends.

\begin{figure}[!t]
\centering
\includegraphics[width=2.5in]{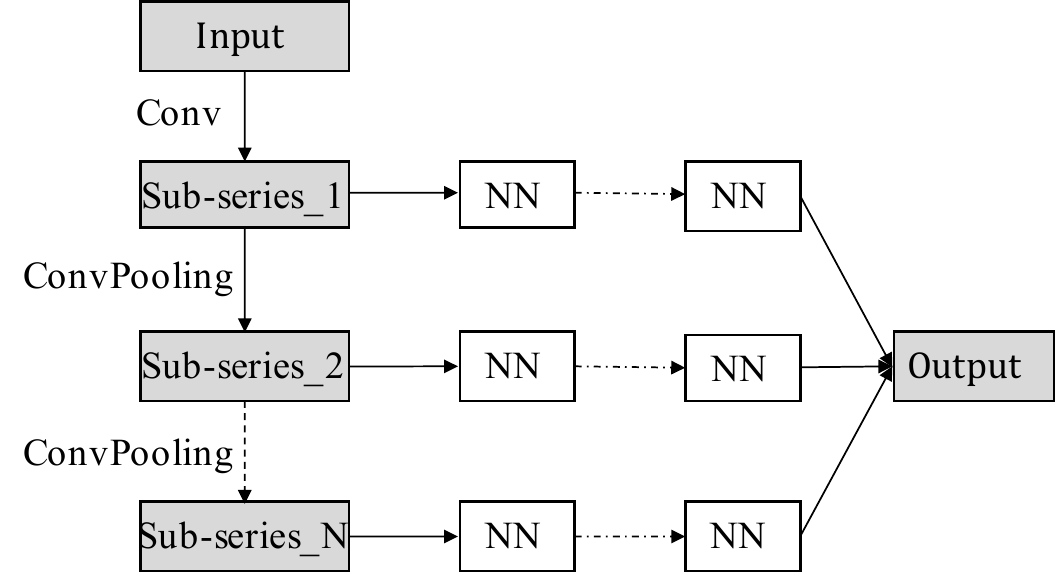}
\caption{The parallel multi-scale structure of SNAS4MTF.}
\label{Figure_2}
\end{figure}

Figure \ref{Figure_2} gives the parallel multi-scale structure of SNAS4MTF. Comparing with the traditional stack-based structure \cite{ref4,ref14}, the advantage of the parallel multi-scale structure is that it can decrease the depth of networks and can alleviate the gradient vanishing problem to some extent.

Our multi-scale decomposition network generates multi-scale sub-series through decomposition layers. Specifically, given the input MTS $\boldsymbol{X}\in\mathbb{R}^{N{\times}T}$, the multi-scale decomposition network generates the sub-series of $K$ scales, and the $k^\text{th}$ scale sub-series is denoted as $\boldsymbol{X}^k\in\mathbb{R}^{N\times{\frac{T}{2^{k-1}}\times{c^k}}}$, where $N$ is the variable dimension, $\frac{T}{2^{k-1}}$  is the sequence length in the $k^\text{th}$ scale, and $c^k$ is the feature dimension of the $k^\text{th}$ scale.

The decomposition layers use convolution and pooling operations to effectively capture local patterns in the time dimension. In the $1^\text{th}$ decomposition layer, a convolution operation is used to obtain the output of the $1^\text{th}$ scale sub-series:
\begin{equation}
\label{5}
\boldsymbol{X}^1={ReLU}\left(\boldsymbol{W}^1\otimes\boldsymbol{X}+\boldsymbol{b}^1\right),
\end{equation}
where $\otimes$ denotes convolution operation, $\boldsymbol{W}^1$ and $\boldsymbol{b}^1$ denote the convolution kernel and bias vector in the $1^\text{th}$ decomposition layer, respectively. In the subsequent decomposition layers, convolution and pooling operations are used. The $k^\text{th}$ decomposition layer can be represented as:
\begin{equation}
\label{6}
\boldsymbol{X}^k={Poling}\left({ReLU}\left(\boldsymbol{W}^k\otimes\boldsymbol{X}^{k-1}+\boldsymbol{b}^k\right)\right),
\end{equation}
where $\boldsymbol{W}^k$ and $\boldsymbol{b}^k$ denote the convolution kernel and bias vector in the $k^\text{th}$ decomposition layer, respectively.

After that, the obtained multi-scale sub-series are flexible and comprehensive to preserve various kinds of temporal dependencies. Since the inter-variable dependencies will be explicitly modeled by GNN, the convolution operations are only performed on the time dimension and the variable dimension is fixed to avoid the interaction between the variables. The kernels are shared across all variables at each decomposition layer.
\subsection{Adaptive Graph Learning}
The adaptive graph learning module automatically generates adjacency matrices to represent the inter-variable dependencies among MTS. Existing learning-based methods \cite{ref12,ref13} only learn a shared adjacency matrix, which is useful to learn the most prominent inter-variable dependencies among MTS in many problems, and can significantly reduce the number of parameters and avoid the overfitting problem. However, the inter-variable dependencies may be different under different time scales. The shared adjacency matrix makes the models biased to learn one type of prominent and shared temporal patterns. Therefore, it is essential to learn multiple scale-specific adjacency matrices.

A straightforward strategy is to learn an adjacency matrix for each scale, which will introduce too many parameters and make the model hard to train. To solve this problem, we propose an adaptive graph learning (AGL) module, which shares partial parameters between learning multiple adjacency matrices. AGL consists of two sub-modules: a scale shared graph learner and scale-specific graph learners. The scale shared graph learner generates a basic adjacency matrix, which can reflect the inherent characteristics of the dataset. The scale-specific graph learners take the basic adjacency matrix as input and outputs multiple scale-specific adjacency matrices, which can reflect all kinds of scale-specific inter-variable dependencies hidden in the dataset. In addition, to further reduce the number of parameters, matrix decomposition is introduced to the scale shared graph learner. Specifically, we initialize the node embeddings $\boldsymbol{E}_\text{nodes}\in\mathbb{R}^{N\times{d_\text{e}}}$ (where $d_\text{e}$ is the embedding dimension and $d_\text{e}\ll N$) and use matrix manipulation to obtain the basic adjacency matrix, rather than learn an adjacency matrix $\boldsymbol{A}_{\text{basic}}\in\mathbb{R}^{N\times N}$ straightforwardly. Figure 3 gives the details of AGL. The scale shared graph learner takes $\boldsymbol{E}_\text{nodes}$ into an MLP layer:
\begin{equation}
\label{7}
\boldsymbol{E}^{\prime}_{\text{nodes}}={MLP}\left(\boldsymbol{E}_\text{nodes}\right).
\end{equation}

Then, the pairwise node similarities are calculated to construct the basic adjacency matrix \cite{ref4}:
\begin{equation}
\label{8}
\begin{aligned}
&\boldsymbol{M}_1=\left[tanh\left(\boldsymbol{E}^{\prime}_{\text{nodes}}\boldsymbol{\theta}\right)\right]^T,\\
&\boldsymbol{M}_2=tanh\left(\boldsymbol{E}^{\prime}_{\text{nodes}}\boldsymbol{\varphi}\right),\\
&\boldsymbol{A}_{\text{basic}}=MLP\left(ReLU\left(\boldsymbol{M}_1\boldsymbol{M}_2-\left(\boldsymbol{M}_2\right)^T\left(\boldsymbol{M}_1\right)^T\right)\right),
\end{aligned}
\end{equation}
where $\boldsymbol{\theta}\in\mathbb{R}^{1\times{1}}$ and $\boldsymbol{\varphi}\in\mathbb{R}^{1\times{1}}$ are learnable parameters. $\boldsymbol{A}_{\text{basic}}\in\mathbb{R}^{N\times{N}}$ denotes the basic adjacency matrix.

The scale-specific graph learners take $\boldsymbol{A}_{\text{basic}}$ as input and output multiple scale-specific adjacency matrices. For the $k^{\text{th}}$ scale-specific graph learner, i.e., the scale-specific graph learner of the $k^{\text{th}}$ scale, $\boldsymbol{A}_{\text{basic}}$ is fed into an MLP layer to learn the scale-specific adjacency matrix:
\begin{equation}
\label{9}
\boldsymbol{A}^{k}_{\text{scale}}={MLP}^k\left(\boldsymbol{A}_\text{basic}\right),
\end{equation}
where $\boldsymbol{A}^{k}_{\text{scale}}\in\mathbb{R}^{N\times{N}}$ is the $k^{\text{th}}$ scale-specific adjacency matrix, and the values in $\boldsymbol{A}_{\text{scale}}^k\in\mathbb{R}^{N\times{N}}$ are normalized to $\left[0,1\right]$, which are regarded as the soft edges among the nodes. To make the model more robust, reduce the computation cost of the graph convolution, and reduce the impact of noise, we introduce a strategy to make $\boldsymbol{A}_{\text{scale}}^k$ sparse:
\begin{equation}
\label{10}
\boldsymbol{A}^{k}={Sparse}\left({Softmax}\left(\boldsymbol{A}_\text{scale}^k\right)\right),
\end{equation}
where $\boldsymbol{A}^k\in\mathbb{R}^{N\times{N}}$ is the final $k^{\text{th}}$ scale-specific adjacency matrix, ${Softmax}$ function is used to normalization, and ${Sparse}$ function is defined as:
\begin{equation}
\label{11}
\boldsymbol{A}_{ij}^k=
\begin{cases}\boldsymbol{A}_{ij}^k,\qquad\boldsymbol{A}_{ij}^k\in TopK\left(\boldsymbol{A}_{i*}^k,\tau\right)\\0,\quad\qquad\boldsymbol{A}_{ij}^k\notin TopK\left(\boldsymbol{A}_{i*}^k,\tau\right)
\end{cases},
\end{equation}
where $\tau$ is the threshold of $TopK$ function and denotes the max number of neighbors of a node.
\begin{figure*}[!t]
\centering
\includegraphics[width=5in]{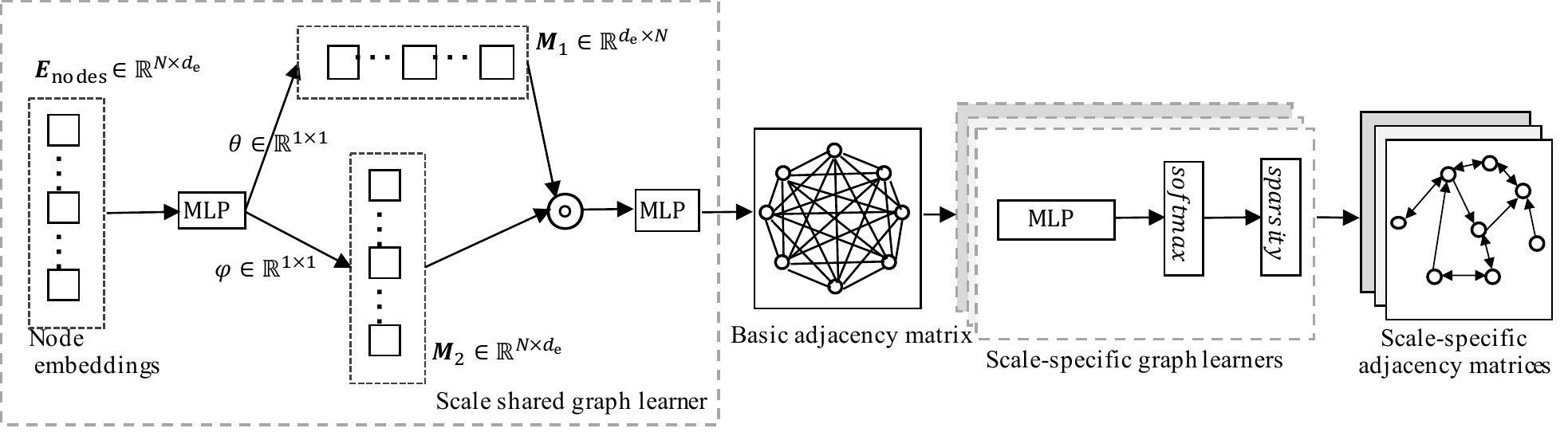}
\caption{The detailed architecture of the adaptive graph learning module.}
\label{Figure 3.}
\end{figure*}
\subsection{Search Space}
The search space in NAS defines a set of operations (e.g., convolution, pooling, and fully-connected) and how these operations can be connected to form valid network architectures. Following DARTS framework \cite{ref21}, SNAS4MTF employs cells as building blocks, and builds a network for each scale by stacking a cell with a predefined number of times. SNAS4MTF searches an optimal cell architecture for each scale.

{\bf{Cells.}} As shown in Figure \ref{Figure_1}, a cell is a directed acyclic graph, where each node $\boldsymbol{\mathcal{X}}^i$ represents a latent representation, and the connection $\left(i,j\right)$ represents a set of operations $\mathcal{O}^{\left(i,j\right)}=\{\sigma_1^{\left(i,j\right)},\sigma_2^{\left(i,j\right)},...,\sigma_\mathcal{K}^{\left(i,j\right)}\}$ that transform $\boldsymbol{\mathcal{X}}^i$, where $\mathcal K$ denotes the number of operations. The output of a cell is obtained by applying an add operation on all the intermediate nodes.

The output $\boldsymbol{\mathcal{C}}^{\left(i,j\right)}$ of a connection $\left(i,j\right)$ is the weighted sum of the outputs of all candidate operations:
\begin{equation}
\label{12}
\boldsymbol{\mathcal{C}}^{\left(i,j\right)}=\sum\nolimits_{k=1}^\mathcal{K}\alpha_k^{(i,j)}\sigma_k^{(i,j)}\left(\boldsymbol{\mathcal{X}}^i\right),
\end{equation}
where $\sigma_k^{(i,j)}\left(\boldsymbol{\mathcal{X}}^i\right)$ denotes the output of the $k^\text{th}$ operation, $\alpha^{(i,j)}=\{\alpha_1^{(i,j)},\alpha_2^{(i,j)},...,\alpha_\mathcal{K}^{(i,j)}\}$ is the architecture parameter of the connection $\left(i,j\right)$, and $\alpha_k^{(i,j)}$ denotes the probability of selecting $\sigma_k^{(i,j)}$.

The representation $\boldsymbol{\mathcal{X}}^j$ of node $j$ is the sum of all the incoming connections of this node:
\begin{equation}
\label{13}
\boldsymbol{\mathcal{X}}^j=\sum\nolimits_{i<j}\boldsymbol{\mathcal {C}}^{\left(i,j\right)}.
\end{equation}

Initially, the cells have undetermined architectures, and the goal is to search an optimal architecture for each of them, i.e., for each connection $\left(i,j\right)$, the operation with highest probability in $\alpha^{(i,j)}$ is selected as the operation in the final architecture.

{\bf{Candidate operations.}} The design of candidate operations is the most important part of the search space. For MTS forecasting, the operations in the search space should effectively model both intra-variable dependencies and inter-variable dependencies. Following the existing design philosophy of NAS works and MTS forecasting works \cite{ref4,ref19,ref21}, we divide candidate operations into three groups: basic operations, the operations for capturing intra-variable dependencies, and the operations for capturing inter-variable dependencies.

The basic operations include Zero operation, Identity operation, and Conv\_1 operation. Zero operation outputs a vector with the same size as the input vector and the values of the output vector are 0. Identity operation outputs the same vector as the input vector. Conv\_1 operation is a standard convolution with kernel size $1\times 1$. These basic operations enable multiple branches and shortcut connections in network architectures.

The operations for capturing intra-variable dependencies include temporal convolution (T-Conv) and temporal attention (T-Att).

T-Conv is used to capture intra-variable dependencies by performing convolution operations on the time dimension for each time series. Given the input feature $\boldsymbol{\mathcal{X}}=\left[\boldsymbol{x}_1,\boldsymbol{x}_2,…,\boldsymbol{x}_N\right]\in\mathbb{R}^{N\times T_l\times c_l}$, where $\boldsymbol{x}_i\in\mathbb{R}^{T_l\times c_l}$ represents the feature of the $i^{\text{th}}$ time series, $T_l$ and $c_l$ are the sequence length and feature dimension in the $l^{\text{th}}$ layer of a cell, respectively. For $\boldsymbol{x}_i$, T-Conv performs two convolution operations and their outputs are bitwise added:
\begin{equation}
\label{14}
\ \boldsymbol{x}^{\prime}_i=Tanh\left(\boldsymbol{W}_1\otimes \boldsymbol{x}_i+\boldsymbol{b}_1\right)\ast{Sigmoid}\left(\boldsymbol{W}_2\otimes \boldsymbol{x}_i+\boldsymbol{b}_2\right),
\end{equation}
where $\boldsymbol{W}_*$ and $\boldsymbol{b}_*$ denote the convolution kernel and bias vector. One convolution operation is followed by a ${Tanh}$ activation function that works as a filter. The other convolution operation is followed by a ${Sigmoid}$ activation function and works as a gate to control the amount of information the filter can pass to the next operation.

T-Conv is performed on the time dimension and the kernels are shared between the variable dimension. The output of T-Conv is $\boldsymbol{\mathcal{X}}^{\prime}=\{\boldsymbol{x}_1^{\prime},\boldsymbol{x}_2^{\prime},…,\boldsymbol{x}_N^{\prime}\}\in\mathbb{R}^{N\times T_l\times c_l}$.

T-Att is used to capture intra-variable dependencies by utilizing temporal attention mechanism. Given the input feature $\boldsymbol{\mathcal{X}}\in\mathbb{R}^{N\times{T_l}\times c_l}$, T-Att calculates a temporal attention matrix to represent the importance of each time step:
\begin{equation}
\label{15}
\boldsymbol{\mathbi{I}}=\boldsymbol{\mathbi{V}}_e\ast{Sigmoid}\left(\left(\boldsymbol{\mathcal{X}}\mathbi{U}_1\right){\mathbi{U}}_2\left({\mathbi{U}}_3\boldsymbol{\mathcal{X}}^T\right)+\mathbi{b}_\text{e}\right),
\end{equation}
where $\mathbi{V}_\text{e}\in\mathbb{R}^{T_l\times T_l}, \mathbi{b}_\text{e}\in\mathbb{R}^{T_l\times T_l}, \mathbi{U}_1\in\mathbb{R}^N, \mathbi{U}_2\in\mathbb{R}^{c_l\times N}$, and $\mathbi{U}_3\in\mathbb{R}^{c_l}$ are learnable parameters. $\mathbi{I}_{i,j}$ indicates the impact of the $j^\text{th}$ time step on the $i^\text{th}$ time step. $\mathbi{I}$ is normalized by ${Softmax}$ function:
\begin{equation}
\label{16}
\mathbi{I}^{\prime}_{i,j}=\frac{\textit{exp}\left(\mathbi{I}_{i,j}\right)}{\sum_{j=0}^{T_{l-1}}\textit{exp}\left(\mathbi{I}_{i,j}\right)}.
\end{equation}

The normalized temporal attention matrix $\mathbi{I}^{\prime}$ is multiplied with input feature $\boldsymbol{\mathcal{X}}$ to obtain the output of T-Att $\boldsymbol{\mathcal{X}}^{\prime}$:
\begin{equation}
\label{17}
\boldsymbol{\mathcal{X}}^{\prime}=\boldsymbol{\mathcal{X}}\mathbi{I}^{\prime}.
\end{equation}

The operations for capturing inter-variable dependencies include graph convolution (GCN) and spatial attention (S-Att).

GCN is used to capture inter-variable dependencies by utilizing graph convolution network. Given the input feature $\boldsymbol{\mathcal{X}}=[\boldsymbol{x}_1,\boldsymbol{x}_2,…,\boldsymbol{x}_{T_l}]\in\mathbb{R}^{N\times T_l\times c_l}$, where $\boldsymbol{x}_i\in\mathbb{R}^{N\times c_l} $ represents the feature at the $i^\text{th}$ time step, and the learned adjacency matrix $\mathbi{A}$, for $\boldsymbol{x}_i$, a graph convolution is performed:
\begin{equation}
\label{18}
\mathbi{x}^{\prime}_i={GCN}\left(\mathbi{x}_i,\mathbi{A},\mathbi{W}\right),
\end{equation}
where $\boldsymbol W$ denotes the learnable parameters of graph convolution. The parameters of graph convolution at different time steps are shared. The output of $\text{GCN}$ is $\boldsymbol{\mathcal{X}}^{\prime}=\{\boldsymbol{x}_1^{\prime},\boldsymbol{x}_2^{\prime},…,\boldsymbol{x}_{T_l}^{\prime}\}\in\mathbb{R}^{N\times T_l\times c_l}$.

S-Att is used to capture inter-variable dependencies by utilizing spatial attention mechanism. Given the input feature $\boldsymbol{\mathcal{X}}\in\mathbb{R}^{N\times T_l\times c_l}$, S-Att calculates a spatial attention matrix to represent the importance of each variable:
\begin{equation}
\label{19}
\mathbi{E}=\mathbi{V}_\text{s}\ast{Sigmoid}\left(\left(\boldsymbol{\mathcal{X}}^T\mathbi{U}_4\right)\mathbi{U}_5\left(\mathbi{U}_6\boldsymbol{\mathcal{X}}\right)+\mathbi{b}_\text{s}\right),
\end{equation}
where $\mathbi{V}_\text{s}\in\mathbb{R}^{N\times N}, \mathbi{b}_\text{s}\in\mathbb{R}^{N\times N}, \mathbi{U}_4\in\mathbb{R}^{T_l}, \mathbi{U}_5\in\mathbb{R}^{c_l\times{T_l}},$ and $\mathbi{U}_6\in\mathbb{R}^{c_l}$ are learnable parameters. $\mathbi{E}_{i,j}$ indicates the impact of the $j^\text{th}$ variable on the $i^\text{th}$ variable. $\mathbi{E}$ is normalized by $Softmax$ function:
\begin{equation}
\label{20}
\mathbi{E}^{\prime}_{i,j}=\frac{\textit{exp}\left(\mathbi{E}_{i,j}\right)}{\sum_{j=0}^{N-1}\textit{exp}\left(\mathbi{E}_{i,j}\right)}.
\end{equation}

The normalized spatial attention matrix $\boldsymbol{E}^{\prime}$ is multiplied with input feature $\boldsymbol{\mathcal{X}}$ to obtain the output of S-Att $\boldsymbol{\mathcal{X}}^{\prime}$:
\begin{equation}
\label{21}
\boldsymbol{\mathcal{X}}^{\prime}=\boldsymbol{\mathcal{X}}\mathbi{E}^{\prime}.
\end{equation}

{\bf{Connections.}} In each connection, selecting an optimal operation from all candidate operations may cause large memory consumption and decrease the convergence speed. To solve the memory inefficient problem, the connections are divided into two groups: adjacent connections (i.e., the black dash lines in Figure \ref{Figure_1}(c)) and skip connections (i.e., the grey dash lines in Figure \ref{Figure_1}(c)). Since adjacent connections are more important for capturing intra-variable dependencies and inter-variable dependencies, we select operations for adjacent connections from the operation pool \{Zero, Identity, Conv\_1, T-Conv, T-Att, GCN, S-Att\}, and we select operations for skip connections from the operation pool \{Zero, Identity, Conv\_1, T-Conv, GCN, which excludes T-Att and S-Att\}. Through selecting operations from different operation pools, the number of possible network architectures can be reduced and the training process can be more efficient.

{\bf{Complexity analysis.}} Given the number of nodes in a cell ${M}$, and the number of cells ${C}$, when connections are not grouped, there are ${M}(M-1)/2$ connections in a cell and each connection contains 7 candidate operations. i.e., there are $7^{{M({M-1})/{2}}}$ possible network architectures in a cell and there are $7^{{M(M-1)C}/{2}}$ possible network architectures in total. When connections are grouped, there are $(M-1)$ adjacent connections and ${(M-2)(M-1)}/{2}$ skip connections in a cell. The number of possible network architectures is $7^{(M-1)C}+5^{{(M-2)(M-1)C}/{2}}$.
\subsection{Fusion Module \& Objection Function}
For a given sub-series, we can stack one or more cells to extract the scale-specific features, and the output of the last cell is the representation of the sub-series. Then, we concatenate these scale-specific representations and obtain the final multi-scale representation. Finally, a two-layer MLP network is used to transform the multi-scale representation into the desired output dimension and obtain the predicted values $\left[\boldsymbol{\hat{x}}_1,\boldsymbol{\hat{x}}_2,...,\boldsymbol{\hat{x}}_h\right]\in\mathbb{R}^{N\times h}$.

We choose $\mathcal{L}_2$ loss as the objection function of our forecasting task, which can be formulated as:
\begin{equation}
\label{22}
\mathcal{L}=\frac{1}{\mathcal{T}_{\text{train}}}\sum\nolimits_{i=1}^{\mathcal{T}_{\text{train}}}\sum\nolimits_{t=1}^h\sum\nolimits_{j=1}^N\left(\boldsymbol{\widehat x}_{t,j}\left(i\right)-\boldsymbol{x}_{t,j}\left(i\right)\right)^2,
\end{equation}
where $\mathcal{T}_{\text{train}}$ is the number of training samples, $N$ is the number of variables. $\boldsymbol{\widehat{x}}_{t,j}\left(i\right)$ and $\boldsymbol{x}_{t,j}\left(i\right)$ are the predicted value and ground true of the $j^\text{th}$ variable at time step $t$ in the $i^\text{th}$ sample, respectively.
\subsection{Training}
Following DARTS framework \cite{ref21}, to make the search space continuous, the categorical choices of the operations are relaxed by learning the continuous variables architecture parameters $\theta_\text{a}=\{\alpha^{(i,j)}\}$. The training of SNAS4MTF contains two stages: search stage and train stage. Algorithm 1 gives the training process of SNAS4MTF. First, dataset is divided into training dataset $\mathcal{D}_{\text{train}}$ and validation dataset $\mathcal{D}_\text{valid}$. During the search stage, the weight parameters of networks (e.g., the weights of the convolution filters) and architecture parameters are optimized alternatively. The training loss $\mathcal{L}_2{\text{(train)}}$ is exploited to optimize the weight parameters $\theta_\text{w}$. The validation loss $\mathcal{L}_2{\text{(valid)}}$ is exploited to optimize the architecture parameters $\theta_\text{a}$. We can get the optimized architecture parameters $\theta_\text{a}^{\ast}$ when the number of iterations is reached. During the train stage, we fix the architecture by choosing the networks with the maximum values in $\theta_\text{a}^{\ast}$. Then, the training loss $\mathcal{L}_2\text{(train)}$ is exploited to optimize the weight parameters $\theta_\text{w}$ and we can obtain the optimized SNAS4MTF model.
\begin{algorithm}[H]
\caption{The training process of SNAS4MTF.}\label{alg:alg1}
\begin{algorithmic}
\STATE 
\STATE {\text{\quad Input:}} 
\STATE {\text{\quad Training dataset}} $\mathcal{D}_\text{train},${\text{validation dataset}} $\mathcal{D}_\text{valid}$
\STATE {\text{\quad Initialized architecture parameters:}} $\theta_\text{a}$
\STATE {\text{\quad Initialized weight parameters:}} $\theta_\text{w}$
\STATE {\text{\quad Adjustable learning rate:}}$\eta_1$ {\text{and}} $\eta_2$
\STATE {\text{\quad Output:}}
\STATE {\text{\quad Learned SNAS4MTF model (optimized }}$\theta_\text{a} {\text{ and }}\theta_\text{w}${\text{)}}
\STATE {\text{\quad\# The search stage of SNAS4MTF}}
\STATE {\text{1\quad{\textbf{Repeat}}}}
\STATE {\text{ \quad\qquad\#  update weight parameters}}
\STATE {\text{2\quad\qquad Random sample a batch from }}$\mathcal{D}_{\text{train}}$
\STATE {\text{3\quad\qquad Calculate loss $\mathcal{L}_2\text{(train)}$ based on the sampled batch}}
\STATE {\text{4\quad\qquad Update weight parameters: }}$\theta_\text{w}\gets\theta_\text{w}-\eta_1\frac{\partial\mathcal{L}_2{\text{(train)}}}{\partial\theta_{\text{w}}}$
\STATE {\text{ \quad\qquad\#  update architecture parameters}}
\STATE {\text{5\quad\qquad Random sample a batch from}} $\mathcal{D}_{\text{valid}}$
\STATE {\text{6\quad\qquad Calculate loss $\mathcal{L}_2 {\text{(valid)}}$ based on the sampled batch}}
\STATE {\text{7\quad\qquad Update weight parameters: }}$\theta_\text{a}\gets\theta_\text{a}-\eta_2\frac{\partial\mathcal{L}_2{\text{(valid)}}}{\partial\theta_{\text{a}}}$
\STATE {\text{8\quad\textbf{Until} reach the number of iterations}}
\STATE {\text{9\quad Get the optimized architecture parameters}} $\theta_\text{a}^{\ast}$
\STATE {\text{ \quad\# The train stage of SNAS4MTF}}
\STATE {\text{10\quad Fix the architecture according to}} $\theta_\text{a}^{\ast}$
\STATE {\text{11\quad Initialize the weight parameters}} $\theta_{\text{w}}$
\STATE {\text{12\quad \textbf{Repeat}}}
\STATE {\text{13\quad\qquad Random sample a batch from }}$\mathcal{D}_\text{train}$
\STATE {\text{14\quad\qquad Calculate loss $\mathcal{L}_2 \text{(train)}$ based on the sampled batch}}
\STATE {\text{15\quad\qquad Update weight parameters: }}$\theta_\text{w}\gets\theta_\text{w}-\eta_1\frac{\partial\mathcal{L}_2{\text{(train)}}}{\partial\theta_{\text{w}}}$
\STATE {\text{16\quad\textbf{Until} reach the number of iterations}}
\STATE {\text{17\quad\textbf{Return} learned SNAS4MTF model}}
\end{algorithmic}
\label{alg1}
\end{algorithm}
\section{Experiments}
\subsection{Datasets and Settings}
To evaluate the performance of SNAS4MTF, we conduct experiments on two public benchmark datasets: METR-LA and PEMS-BAY. The details of these datasets are given as follows.

\begin{itemize}
\item{{\bf{METR-LA:}} This dataset is collected by the Los Angeles Metropolitan Transportation Authority and contains the average traffic speed measured by 207 loop detectors on the highways of Los Angeles County between March 2012 and June 2012.}
\item{{\bf{PEMS-BAY:}} This dataset is collected by California Transportation Agencies and contains the average traffic speed measured by 325 sensors in the Bay Area between January 2017 and May 2017.}
\end{itemize}

Following existing works \cite{ref4,ref13,ref19}, the two datasets are split into the training set (70\%), validation set (20\%), and test set (10\%) in chronological order.

For experimental settings, following existing works \cite{ref4, ref13, ref19}, the length of input time series is 12, and the forecasting values contain the next 12 future steps, i.e., the traffic flow within one hour in the future is predicted with a time interval of 5 minutes. The time of the day is used as an auxiliary feature for the inputs. The number of scales $K$ in the multi-scale decomposition network is 3. The number of neighbors $\tau$ is set as 20. The node embedding dimension $d_\text{e}$ in the adaptive graph learning is 20. The hidden dimension $d_\text{h}$ of the operations is 32. The number of cells in 3 scales is 1, 2, and 2, respectively. Each cell has 4 nodes. The number of epochs in the search stage and the train stage are set to 60 and 100, respectively. The initial learning rates for weight parameters and architecture parameters are set to 0.01 and 0.001, respectively.

Mean Absolute Error (MAE), Mean Absolute Percentage Error (MAPE), and Root Mean Squared Error (RMSE) are exploited as evaluation metrics. For the three metrics, the smaller the value is, the better the performance of the model is. We implement our model with PyTorch \cite{ref31} and construct GNNs with PyTorch Geometric library \cite{ref32}. The sourse code is released on GitHub\footnote{https://github.com/shangzongjiang/SNAS4MTF}. We implement training, validation, and test tasks on a server with one GPU (NVIDIA RTX 2080Ti).
\subsection{Methods for Comparison}
The models in our comparative evaluation are as follows.

\noindent{\bf{Spatio-temporal methods requiring a pre-specified graph, which is constructed through the road network:}}
\begin{itemize}
\item{DCRNN \cite{ref15}: A diffusion convolutional recurrent neural network, which combines diffusion convolution operations and sequence-to-sequence architecture.}
\item{STGCN \cite{ref16}: A spatio-temporal graph convolutional network, which integrates graph convolution operations with gated temporal convolution operations.}
\item{Graph WaveNet \cite{ref13}: It exploits graph convolution operations and WaveNet to model spatial and temporal correlations, respectively.}
\item{ST-MetaNet \cite{ref39}: A meta learning based spatial-temporal network, which consists two types of meta networks: meta graph attention network and meta recurrent neural network.}
\item{GMAN \cite{ref14}: A graph multi-attention network, which integrates multiple spatio-temporal attention blocks, and a transform attention layer into the encoder-decoder architecture.}
\item{MRA-BGCN \cite{ref11}: A multi-range attentive bicomponent graph convolution network, which models the interactions of both nodes and edges using bi-component graph convolution.}
\item{AutoSTG \cite{ref19}: A spatio-temporal graph forecasting method, which integrates neural architecture search to capture complex spatio-temporal correlations and the meta learning to generate parameters.}
\end{itemize}
{\bf{Pure MTS forecasting methods without a pre-specified graph:}}
\begin{itemize}
\item{ARIMA: A autoregressive integrated moving average method.}
\item{SVR: A support vector regression method.}
\item{FNN: A feedforward neural network.}
\item{FC-LSTM \cite{ref40}: A sequence-to-sequence method that uses fully connected LSTMs in encoder and decoder.}
\item{Graph WaveNet* \cite{ref13}: It exploits graph convolution operations and WaveNet to model spatial and temporal correlations, respectively. The adjacency matrix of the graph convolution is learned from data.}
\item{FC-GAGA \cite{ref12}: A fully connected gated graph architecture, which combines a fully-connected time series model N-BEATS and a hard graph gate mechanism.}
\item{MTGNN \cite{ref4}: A graphing learning based MTS forecasting method, which exploits a graph learning module to learn an adjacency matrix, and models MTS using GNN and dilated convolution.}
\item{SNAS4MTS: It is our proposed method.}
\end{itemize}

For all the baselines, we use the reported results from the original papers, which are published in the top data mining and AI conferences.
\subsection{Main Results}
Tables \ref{tab:table1} and \ref{tab:table2} report the empirical results of all the methods for 15 minutes, 30 minutes, and 1 hour ahead forecasting on the two datasets, respectively. The best performance over the pure MTS forecasting methods is highlighted in bold, and the following tendencies can be discerned:

1) SNAS4MTS compares favourably even against spatio-temporal models that requires a pre-specified graph (DCRNN, STGCN, Graph WaveNet, ST-MetaNet, GMAN, MRA-BGCN, and AutoSTG). SNAS4MTS uses pure MTS data without any prior knowledge, and introduces an adaptive graph learning module to generate the weighted adjacency matrices. The results imply that the adaptive graph learning module can effectively infer the inter-variable dependencies under different time scales.

2) AutoSTG get worse performance than SNAS4MTS. Although AutoSTG incorporates NAS to search the optimal architecture for spatio-temporal graphs. AutoSTG exploits additional information (node attributes and edge attributes) to obtain the adjacency matrix, which is not suitable for pure MTS forecasting. In addition, AutoSTG constructs the network architecture by stacking multiple identical spatio-temporal modules, which makes it fail to model multi-scale patterns explicitly.

3) Traditional MTS forecasting methods (ARIMA and SVR) get worse results than deep learning methods, as they cannot capture the non-stationary and non-linear dependencies.

4) FNN and FC-LSTM are univariate models that cannot model inter-variable dependencies. Thus, they get worse performance than other deep learning based MTS forecasting methods.

5) SNAS4MTS get better performance than graph learning based methods (Graph WaveNet$\ast$, FC-GAGA, and MTGNN). Existing graph learning based methods only learn a shared adjacency matrix, which cannot consider the different inter-variable dependencies under different time scales, resulting in sub-optimal to model multi-scale temporal patterns.
\begin{table*}[]
\centering
\caption{Results summary of all methods on METR-LA dataset.\label{tab:table1}}
\begin{tabular}{llllllllll}
\hline
\multirow{2}{*}{Methods} & \multicolumn{3}{l}{15 min} & \multicolumn{3}{l}{30 min} & \multicolumn{3}{l}{60 min} \\ \cline{2-10} 
                         & MAE    & MAPE     & RMSE   & MAE    & MAPE     & RMSE   & MAE    & MAPE     & RMSE   \\ \hline
DCRNN                    & 2.77   & 7.30\%   & 5.38   & 3.15   & 8.80\%   & 6.45   & 3.60   & 10.50\%  & 7.60   \\ 
STGCN                    & 2.88   & 7.62\%   & 5.74   & 3.47   & 9.57\%   & 7.24   & 4.59   & 12.70\%  & 9.40   \\ 
Graph WaveNet            & 2.69   & 6.90\%   & 5.15   & 3.07   & 8.37\%   & 6.22   & 3.53   & 10.01\%  & 7.37   \\ 
ST-MetaNet               & 2.69   & 6.91\%   & 5.17   & 3.10   & 8.57\%   & 6.28   & 3.59   & 10.63\%  & 7.52   \\ 
GMAN                     & 2.77   & 7.25\%   & 5.48   & 3.07   & 8.35\%   & 6.34   & 3.40   & 9.72\%   & 7.21   \\ 
MRA-BGCN                 & 2.67   & 6.80\%   & 5.12   & 3.06   & 8.30\%   & 6.17   & 3.49   & 10.00\%  & 7.30   \\ 
AutoSTG                  & 2.70   & /        & 5.16   & 3.06   & /        & 6.17   & 3.47   & /        & 7.27   \\ \hline
ARIMA                    & 3.99   & 9.60\%   & 8.21   & 5.15   & 12.70\%  & 10.45  & 6.90   & 17.40\%  & 13.23  \\
SVR                      & 3.99   & 9.30\%   & 8.45   & 5.05   & 12.10\%  & 10.87  & 6.72   & 16.70\%  & 13.76  \\
FNN                      & 3.99   & 9.90\%   & 7.94   & 4.23   & 12.90\%  & 8.17   & 4.49   & 14.00\%  & 8.69   \\
FC-LSTM                  & 3.44   & 9.60\%   & 6.3    & 3.77   & 10.90\%  & 7.23   & 4.37   & 13.20\%  & 8.69   \\
Graph WaveNet*           & 2.80   & 7.45\%   & 5.45   & 3.18   & 9.00\%   & 6.42   & 3.57   & 10.47\%  & 7.29   \\
FC-GAGA                  & 2.75   & 7.25\%   & 5.34   & 3.10   & 8.57\%   & 6.30   & 3.51   & 10.14\%  & 7.31   \\
MTGNN                    & 2.69   & \textbf{6.86\%}   & 5.18   & 3.05   & \textbf{8.19\%}   & 6.17   & 3.49   & 9.87\%   & 7.23   \\
SNAS4MTF                 & $\textbf{2.68}$   & 6.90\%   & $\textbf{5.16}$   & \textbf{3.03}   & 8.25\%   & \textbf{6.12}   & \textbf{3.43}   & \textbf{9.82\%}   & \textbf{7.15}   \\ \hline
\end{tabular}
\end{table*}

\begin{table*}[]
\centering
\caption{Results summary of all methods on PEMS-BAY dataset.\label{tab:table2}}
\begin{tabular}{llllllllll}
\hline
\multirow{2}{*}{Methods} & \multicolumn{3}{l}{15 min} & \multicolumn{3}{l}{30 min} & \multicolumn{3}{l}{60 min} \\ \cline{2-10} 
                         & MAE    & MAPE     & RMSE   & MAE    & MAPE     & RMSE   & MAE    & MAPE     & RMSE   \\ \hline
DCRNN                    & 1.38   & 2.90\%   & 2.95   & 1.74   & 3.90\%   & 3.97   & 2.07   & 4.90\%   & 4.74   \\ 
STGCN                    & 1.36   & 2.90\%   & 2.96   & 1.81   & 4.17\%   & 4.27   & 2.49   & 5.79\%   & 5.69   \\ 
Graph WaveNet            & 1.30   & 2.73\%   & 2.74   & 1.63   & 3.67\%   & 3.70   & 1.95   & 4.63\%   & 4.52   \\ 
ST-MetaNet               & 1.36   & 2.82\%   & 2.9    & 1.76   & 4.00\%   & 4.02   & 2.20   & 5.45\%   & 5.06   \\ 
GMAN                     & 1.34   & 2.81\%   & 2.82   & 1.62   & 3.63\%   & 3.72   & 1.86   & 4.31\%   & 4.32   \\ 
MRA-BGCN                 & 1.29   & 2.90\%   & 2.72   & 1.61   & 3.80\%   & 3.67   & 1.91   & 4.60\%   & 4.46   \\ 
AutoSTG                  & 1.31   & /        & 2.76   & 1.63   & /        & 3.67   & 1.92   & /        & 4.38   \\ \hline
ARIMA                    & 1.62   & 3.50\%   & 3.30   & 2.33   & 5.40\%   & 4.76   & 3.38   & 8.30\%   & 6.50   \\
SVR                      & 1.85   & 3.80\%   & 3.59   & 2.48   & 5.50\%   & 5.18   & 3.28   & 8.00\%   & 7.08   \\
FNN                      & 2.20   & 5.19\%   & 4.42   & 2.30   & 5.43\%   & 4.63   & 2.46   & 5.89\%   & 4.98   \\
FC-LSTM                  & 2.05   & 4.80\%   & 4.19   & 2.20   & 5.20\%   & 4.55   & 2.37   & 5.70\%   & 4.96   \\
Graph WaveNet*           & 1.34   & 2.79\%   & 2.83   & 1.69   & 3.79\%   & 3.80   & 2.00   & 4.73\%   & 4.54   \\
FC-GAGA                  & 1.36   & 2.87\%   & 2.86   & 1.68   & 3.80\%   & 3.80   & 1.97   & 4.67\%   & 4.52   \\
MTGNN                    & 1.32   & 2.77\%   & 2.79   & 1.65   & 3.69\%   & 3.74   & 1.94   & 4.53\%   & 4.49   \\
SNAS4MTF                 & \textbf{1.30}   & \textbf{2.72\%}   & \textbf{2.77}   & \textbf{1.61}   & \textbf{3.63\%}   & \textbf{3.68}   & \textbf{1.89}   & \textbf{4.44\%}   & \textbf{4.36}    \\ \hline
\end{tabular}
\end{table*}

\subsection{Effect of Multi-Scale Decomposition}
To demonstrate the effect of multi-scale decomposition, we evaluate the performance of SNAS4MTF with the following variants.

\begin{itemize}
\item{S-scale-1: It does not decompose the time series into multi-scale sub-series and only learns features from the raw time series.}
\item{S-scale-2: The number of scales in the multi-scale decomposition network is 2.}
\item{S-stack: It learns multi-scale patterns by stacking multiple cells and temporal pooling layers. The cell in S-stack is the same as that in SNAS4MTF, and S-stack searches an optimal network architecture for each time scale. A temporal pooling layer is followed by a cell to extract multi-scale features. Then, the outputs of all cells are concatenated, and a two-layer MLP network is used to obtain the predicted values.}
\end{itemize}

Table \ref{tab:table3} reports the results of these methods for 15 minutes, 30 minutes, and 1 hour ahead forecasting on the two datasets. We can observe that the performance of SNAS4MTF is improved, when the number of scales increases from 1 to 3. This is because more diversified short-term and long-term patterns can be captured by SNAS4MTF. In addition, SNAS4MTF outperforms SNAS4MTF-stack on all the horizons and all the metrics. It indicates that the parallel multi-scale structure is better than the traditional stack-based structure, as the parallel multi-scale structure can alleviate the gradient vanishing problem to some extent.
\begin{table*}[]
\centering
\caption{The results of different decomposition methods.\label{tab:table3}}
\begin{tabular}{lllllllllll}
\hline
\multirow{2}{*}{Datasets} & \multirow{2}{*}{Methods} & \multicolumn{3}{l}{15 min} & \multicolumn{3}{l}{30 min} & \multicolumn{3}{l}{60 min} \\ \cline{3-11} 
                          &                          & MAE    & MAPE     & RMSE   & MAE    & MAPE     & RMSE   & MAE    & MAPE     & RMSE   \\ \hline
\multirow{4}{*}{METR-LA}  & S-scale-1                & 2.75   & 7.12\%   & 5.31   & 3.11   & 8.48\%   & 6.29   & 3.48   & 9.90\%   & 7.22   \\ 
                          & S-scale-2                & 2.73   & 7.09\%   & 5.28   & 3.08   & 8.42\%   & 6.22   & 3.45   & 9.83\%   & 7.12   \\ 
                          & S-stack                  & 2.70   & 7.03\%   & 5.23   & 3.06   & 8.39\%   & 6.22   & 3.46   & 9.92\%   & 7.26   \\ 
                          & SNAS4MTF                 & \textbf{2.68}   & \textbf{6.90\%}   & \textbf{5.16}   & \textbf{3.03}   & \textbf{8.25\%}   & \textbf{6.12}   & \textbf{3.43}   & \textbf{9.82\%}   & \textbf{7.15}   \\ \hline
\multirow{4}{*}{PEMS-BAY} & S-scale-1                & 1.32   & 2.74\%   & 2.79   & 1.64   & 3.66\%   & 3.71   & 1.91   & 4.49\%   & 4.38   \\ 
                          & S-scale-2                & 1.32   & 2.78\%   & 2.81   & 1.65   & 3.72\%   & 3.73   & 1.93   & 4.58\%   & 4.41   \\ 
                          & S-stack                  & 1.31   & 2.73\%   & 2.79   & 1.62   & 3.68\%   & 3.73   & 1.90   & 4.51\%   & 4.42   \\
                          & SNAS4MTF                 & \textbf{1.30}   & \textbf{2.72\%}   &\textbf{2.77}   & \textbf{1.61}   & \textbf{3.63\%}   & \textbf{3.68}   & \textbf{1.89}   & \textbf{4.44\%}   & \textbf{4.36}   \\ \hline
\end{tabular}
\end{table*}
\subsection{Effect of Adaptive Graph Learning}
To investigate the effect of adaptive graph learning, we conduct ablation study by carefully designing the following two variants.

\begin{itemize}
\item{S-shared: It only learns one shared adjacency matrix to represent the inter-variable dependencies under different time scales, i.e., the scale shared graph learner is reserved, and the scale-specific graph learners are replaced by a shared network that has the same structure with the scale-specific graph learners.}
\item{S-non-shared: It learns multiple adjacency matrices without sharing any parameters, i.e., the scale-specific graph learners are reserved, and the scale shared graph learner is replaced by non-shared networks that have the same structure with the scale shared graph learner.}
\end{itemize}

Table \ref{tab:table4} shows the results of these methods for 15 minutes, 30 minutes, and 1 hour ahead forecasting on the two datasets. We can observe that SNAS4MTF obtains the best performance, which implies: 1) S-shared cannot consider the different inter-variable dependencies under different time scales, which makes the model biased to only learn one type of prominent inter-variable dependencies. 2) Although S-non-shared can consider the different inter-variable dependencies under different time scales, it introduces too many parameters and make the model hard to train. Furthermore, the differences between the multiple adjacency matrices in S-non-shared would be large without any constraint, and the dramatic difference between adjacent matrices makes the model difficult to converge to the optimum result.

\begin{table*}[th]
\centering
\caption{The results of different graph learning methods.\label{tab:table4}}
\begin{tabular}{lllllllllll}
\hline
\multirow{2}{*}{Datasets} & \multirow{2}{*}{Methods} & \multicolumn{3}{l}{15 min} & \multicolumn{3}{l}{30 min} & \multicolumn{3}{l}{60 min} \\ \cline{3-11} 
                          &                          & MAE    & MAPE     & RMSE   & MAE    & MAPE     & RMSE   & MAE    & MAPE      & RMSE  \\ \hline
\multirow{3}{*}{METR-LA}  & S-shared                 & 2.70   & 7.01\%   & 5.21   & 3.06   & 8.45\%   & 6.19   & 3.45   & 10.02\%   & 7.16  \\  
                          & S-non-shared             & 2.71   & 6.99\%   & 5.23   & 3.07   & 8.40\%   & 6.20   & 3.48   & 10.01\%   & 7.23  \\ 
                          & SNAS4MTF                 & \textbf{2.68}   & \textbf{6.90\%}   & \textbf{5.16}   & \textbf{3.03}   & \textbf{8.25\%}   & \textbf{6.12}   & \textbf{3.43}   & \textbf{9.82\%}    & \textbf{7.15}  \\ \hline
\multirow{3}{*}{PEMS-BAY} & S-shared                 & 1.31   & 2.75\%   & 2.79   & 1.63   & 3.70\%   & 3.71   & 1.90   & 4.47\%    & 4.36  \\  
                          & S-non-shared             & 1.39   & 2.89\%   & 3.00   & 1.83   & 4.15\%   & 4.21   & 2.36   & 5.85\%    & 5.43  \\  
                          & SNAS4MTF                 & \textbf{1.30}   & \textbf{2.72\%}   & \textbf{2.77}   & \textbf{1.61}   & \textbf{3.63\%}   & \textbf{3.68}   & \textbf{1.89}   & \textbf{4.44\%}    & \textbf{4.36}  \\ \hline
\end{tabular}
\end{table*}

\subsection{Effect of Designed Search Space}
To investigate the effect of designed search space, we conduct ablation study by carefully designing the following four variants.

\begin{itemize}
\item{w/o att: It removes the temporal attention and spatial attention from the search space.}
\item{w/o conv: It removes the temporal convolution and graph convolution from the search space.}
\item{w/o basic: It removes the basic operations (Zero operation, Identity operation, and Conv\_1 operation) from the search space.}
\item{w/o grouping: It removes the strategy that selecting operations from different operation pools. All the connections select operations from all candidate operations.}
\end{itemize}

Table \ref{tab:table5} shows the results of these methods on the two datasets. We can see that: 1) The performance of three variants (w/o att, w/o conv, w/o basic) deteriorates regardless of which operations are removed, suggesting that the designed operations are useful for MTS forecasting. 2) Although the performance of w/o grouping is slightly better than that of SNAS4MTF, w/o grouping cause larger memory consumption, as shown in Table \ref{tab:table6}. Overall, comprehensively considering the slight forecasting performance improvement and the additional memory consumption, SNAS4MTF demonstrates the superiority of selecting operations from different operation pools.

\begin{table*}[th]
\centering
\caption{The results of methods with different search space.\label{tab:table5}}
\begin{tabular}{lllllllllll}
\hline
\multirow{2}{*}{Datasets} & \multirow{2}{*}{Methods} & \multicolumn{3}{l}{15 min} & \multicolumn{3}{l}{30 min} & \multicolumn{3}{l}{60 min} \\ \cline{3-11} 
                          &                          & MAE    & MAPE     & RMSE   & MAE    & MAPE     & RMSE   & MAE    & MAPE      & RMSE  \\ \cline{1-10}
\multirow{5}{*}{METR-LA}  & w/o Att                  & 2.72   & 7.09\%   & 5.26   & 3.07   & 8.43\%   & 6.21   & 3.44   & 9.91\%    & 7.14  \\ 
                          & w/o conv                 & 2.91   & 7.71\%   & 5.68   & 3.41   & 9.67\%   & 6.84   & 4.03   & 12.14\%   & 8.10  \\ 
                          & w/o basic                & 2.71   & 6.98\%   & 5.19   & 3.06   & 8.34\%   & 6.16   & 3.43   & 9.80\%    & 7.08  \\ 
                          & w/o grouping             & \textbf{2.67}   & \textbf{6.87\%}   & \textbf{5.14}   & 3.04   & 8.26\%   & \textbf{6.11}   & 3.44   & 9.90\%    & \textbf{7.13}  \\ 
                          & SNAS4MTF                 & 2.68   & 6.90\%   & 5.16   & \textbf{3.03}   & \textbf{8.25\%}   & 6.12   & \textbf{3.43}   & \textbf{9.82\%}    & 7.15  \\ 
\multirow{5}{*}{PEMS-BAY} & w/o Att                  & 1.30   & 2.72\%   & 2.77   & 1.62   & 3.64\%   & 3.69   & 1.89   & 4.46\%    & 4.37  \\ 
                          & w/o conv                 & 1.37   & 2.86\%   & 2.97   & 1.80   & 4.08\%   & 4.14   & 2.30   & 5.66\%    & 5.29  \\
                          & w/o basic                & 1.38   & 2.89\%   & 2.99   & 1.82   & 4.13\%   & 4.18   & 2.33   & 5.75\%    & 5.36  \\
                          & w/o grouping             & \textbf{1.30}   & \textbf{2.71\%}   & \textbf{2.77}   & \textbf{1.61}   & \textbf{3.61\%}   & \textbf{3.67}   & \textbf{1.89}   & \textbf{4.40\%}    & \textbf{4.36}  \\
                          & SNAS4MTF                 & \textbf{1.30}   & 2.72\%   & \textbf{2.77}   & \textbf{1.61}   & 3.63\%   & 3.68   & \textbf{1.89}   & 4.44\%    & \textbf{4.36}  \\ \hline
\end{tabular}
\end{table*}

\begin{table}[th]
\centering
\caption{The GPU memory usage (GB).\label{tab:table6}}
\begin{tabular}{lcccccc}
\hline
Datasets & w/o grouping & SNAS4MTF \\ \hline
METR-LA  & 9.2          & 8.5      \\ 
PEMS-BAY & 10.7         & 10.2      \\ \hline
\end{tabular}
\end{table}

\subsection{Parameter Study}
\begin{figure}[!t]
\centering
\subfloat[METR-LA dataset]{\includegraphics[width=0.25\textwidth]{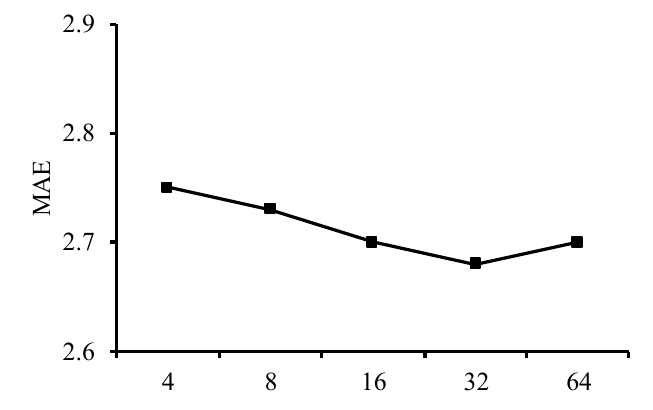}%
\label{fig_first_case}}
\subfloat[PEMS-BAY dataset]{\includegraphics[width=0.25\textwidth]{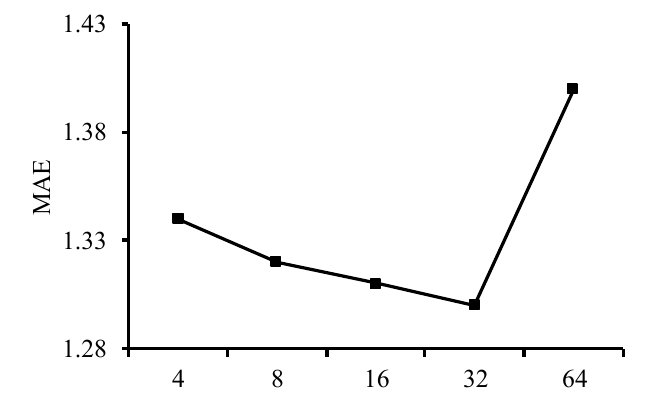}%
\label{fig_second_case}}

\caption{The effect of hidden dimension.}
\label{Figure_4}
\end{figure}

Hidden dimension and node embedding dimension are two important parameters and could influence the performance of SNAS4MTF. Figure \ref{Figure_4} shows the results of SNAS4MTF on these two datasets by varying hidden dimension from 4 to 64. The best performance can be obtained when hidden dimension is 32. It might be that a small hidden dimension limits the expressive ability of SNAS4MTF, and a large hidden dimension would increase the number of parameters and result in the overfitting problem.

\begin{figure}[]
\centering
\subfloat[METR-LA dataset]{\includegraphics[width=0.25\textwidth]{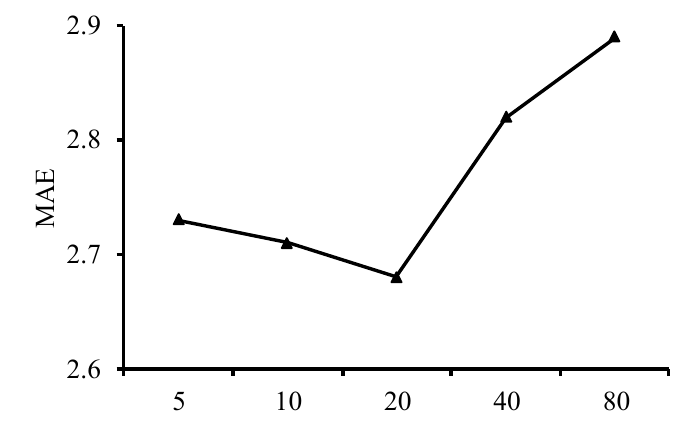}%
\label{fig_first_case}}
\subfloat[PEMS-BAY dataset]{\includegraphics[width=0.25\textwidth]{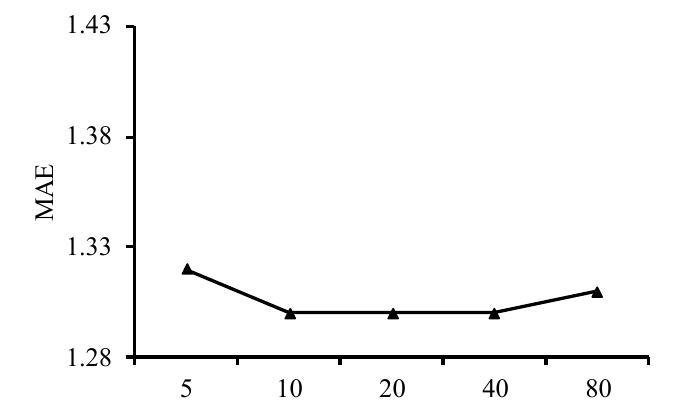}%
\label{fig_second_case}}

\caption{The effect of node embedding dimension.}
\label{Figure_5}
\end{figure}

Figure \ref{Figure_5} shows the results of SNAS4MTF on these two datasets by varying node embedding dimension from 5 to 80. The best performance can be obtained when the node embedding dimension is 20. The node embedding dimension has much larger impact on METR-LA dataset than on PEMS-BAY dataset, as METR-LA is collected in Los Angeles, which is known for its complicated traffic conditions, and METR-LA dataset is more sensitive to hyper-parameters.

\subsection{Case Study}

\begin{figure}[th]
\centering
\subfloat[\tiny{The cell at the $1^\text{th}$ scale on METR-LA dataset}]{\includegraphics[width=1 in]{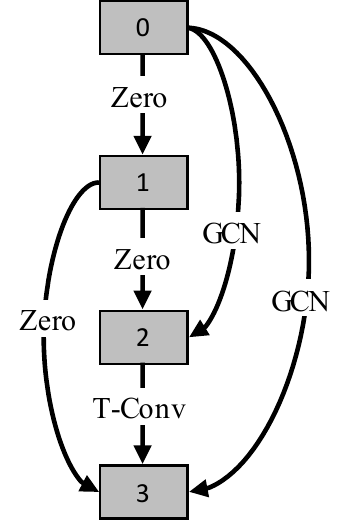}%
\label{Figure6_a}}
\hfil
\subfloat[\tiny{The cell at the $2^\text{th}$ scale on METR-LA dataset}]{\includegraphics[width=1.15in]{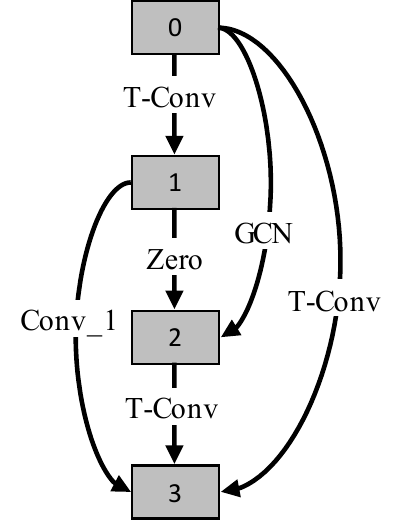}%
\label{Figure6_b}}
\hfil
\subfloat[\tiny{The cell at the $1^\text{th}$ scale on PEMS-BAY dataset}]{\includegraphics[width=1in]{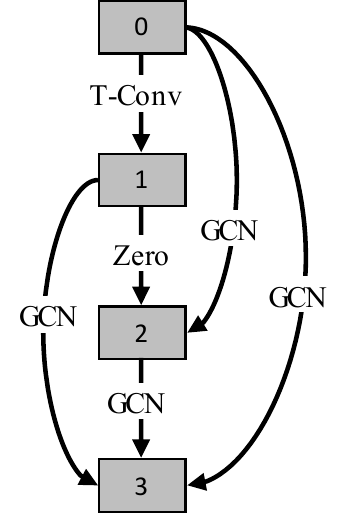}%
\label{Figure6_c}}
\caption{The searched network architectures.}
\label{Figure_6}
\end{figure}

We conduct a case study to evaluate the network architectures searched by SNAS4MTF, and the results are shown in Figure \ref{Figure_6}. Figure \ref{Figure6_a} and Figure \ref{Figure6_b} show the network architecture of the cell at the $1^\text{th}$ scale and the $2^\text{th}$ scale on METR-LA dataset, respectively. We can observe that searched network architectures are different under different time scales. Therefore, it is necessary to search an optimal network architecture for each time scale. Figure \ref{Figure6_c} shows the network architecture of the cell at the $1^\text{th}$ scale on PEMS-BAY dataset. Comparing the network architectures in Figure \ref{Figure6_a} and Figure \ref{Figure6_c}, we can observe that the network architecture searched on METR-LA dataset has more Zero operations, and the network architecture searched on PEMS-BAY dataset has more GCN operations. The reason might be that the neighborhood information is more important for PEMS-BAY dataset. Therefore, SNAS4MTF can automatically find the optimal network architecture for a given scenario.

\section{Conclusions and Future Work}
In this paper, we propose SNAS4MTF to automatically capture both intra-variable dependencies and inter-variable dependencies. The novelty of SNAS4MTF is mainly reflected in two aspects. First, SNAS4MTF designs a search space for MTS forecasting. By decomposing the time series into multi-scale sub-series and searching an optimal network architecture for each time scale, SNAS4MTF can learn multi-scale temporal patterns effectively. Second, SNAS4MTF introduces an adaptive graph learning module, which can infer the different inter-variable dependencies under different time scales. Experimental results on two real-world MTS benchmark datasets have shown that SNAS4MTF outperforms the state-of-the-art baselines.

In the future, this work can be extended in the following directions. First, we will learn dynamic adjacency matrices at different time steps. Second, we will design a mixture graph that describes the inter-variable dependencies and intra-variable dependencies jointly.

\bibliographystyle{unsrt}
\bibliography{SNAS4MTF_20211104V1}

\end{document}